\documentclass[letterpaper, 10 pt, conference]{ieeeconf}
\IEEEoverridecommandlockouts    
\usepackage{subcaption}
\usepackage{tabularx,booktabs}
\PassOptionsToPackage{hyphens}{url}\usepackage{hyperref}
\usepackage{multirow}

\usepackage{graphics}           
\usepackage{times}              
\usepackage{amsmath}            
\usepackage{amssymb}            
\usepackage{graphicx}
\usepackage{algorithm}
\usepackage[noend]{algpseudocode}
\usepackage{booktabs}
\usepackage{color}
\definecolor{instructioncolor}{rgb}{.5,.5,.5}

\usepackage[font=small]{caption}

\def\secref#1{Sec.~\ref{#1}}
\def\figref#1{Fig.~\ref{#1}}
\def\tabref#1{Tab.~\ref{#1}}
\def\eqref#1{Eq.~(\ref{#1})}

\makeatletter
\usepackage{xspace}
\DeclareRobustCommand\onedot{\futurelet\@let@token\@onedot}
\def\@onedot{\ifx\@let@token.\else.\null\fi\xspace}

\def\etal{{et al}\onedot}
\makeatother

\def\etalcite#1{\etal~\cite{#1}}

\usepackage{array}
\newcolumntype{L}[1]{>{\raggedright\let\newline\\\arraybackslash\hspace{0pt}}m{#1}}
\newcolumntype{C}[1]{>{\centering\let\newline\\\arraybackslash\hspace{0pt}}m{#1}}
\newcolumntype{R}[1]{>{\raggedleft\let\newline\\\arraybackslash\hspace{0pt}}m{#1}}

\newcommand{\q}[1]{{{\bf #1}}}

\newcommand{\mq}[1]{{\mbox{{\sffamily{#1}}}}}

\title{\LARGE \bf Fully Onboard Low-Power Localization with Semantic Sensor Fusion on a Nano-UAV using Floor Plans}

\author{Nicky Zimmerman$^{\star}$\dag \and Hanna M\"uller$^{\star}$\ddag \and Michele Magno\ddag \and Luca Benini\ddag \S
  \thanks{$^\star$ Authors contributed equally to this work.}
  \thanks{\dag IDSIA, Universita della Svizzera italiana, Switzerland
  }%
  \thanks{\ddag Integrated Systems Laboratory / Center for Project-Based Learning  - ETH Z\"urich, Switzerland
  }%
  \thanks{\S Department of Electrical, Electronic and Information Engineering - University of Bologna, Italy
  }%
} 

\begin{document}  
\maketitle
\thispagestyle{empty}
\pagestyle{empty}

\begin{abstract}
  %




Nano-sized unmanned aerial vehicles (UAVs) are well-fit for indoor applications and for close proximity to humans. To enable autonomy, the nano-UAV must be able to self-localize in its operating environment. This is a particularly-challenging task due to the limited sensing and compute resources on board. This work presents an online and onboard approach for localization in floor plans annotated with semantic information. Unlike sensor-based maps, floor plans are readily-available, and do not increase the cost and time of deployment. To overcome the difficulty of localizing in sparse maps, the proposed approach fuses geometric information from miniaturized time-of-flight sensors and semantic cues. The semantic information is extracted from images by deploying a state-of-the-art object detection model on a high-performance multi-core microcontroller onboard the drone, consuming only 2.5mJ per frame and executing in 38ms. In our evaluation, we globally localize in a real-world office environment, achieving 90\% success rate. We also release an open-source implementation of our work\footnote{\href{https://github.com/ETH-PBL/Nano-SMCL}{https://github.com/ETH-PBL/Nano-SMCL}}.

\end{abstract}

\vspace{-0.2cm}
\section{Introduction}
\label{sec:intro}


Unmanned aerial vehicles (UAVs) are emerging for applications such as monitoring, inspection, surveillance, transportation and logistics \cite{shakhatreh2019unmanned}. Mainly in indoor scenarios, a small form factor brings key advantages since smaller UAVs allow for safe operation near humans and can reach locations which are inaccessible with larger platforms~\cite{gyagenda2022review}. 

Localization is essential in enabling autonomy for mobile robots. For standard-sized robots, RTK-GPS is often used for outdoor localization~\cite{schneider16icra}, as well as heavy high-end power-hungry 3D LiDARs~\cite{chen2019iros}\cite{elhousni2020iv}, which require expensive computations. 
These approaches are unsuitable for nano-UAVs. First,  nano-UAVs are usually deployed indoors, in GPS-denied environments. Second, their low payload and limited battery capacity restrict the type and accuracy of sensors and the computational resources available onboard.  

One approach to localizing small-sized UAVs in GPS-denied environment is infrastructure-based localization, commonly implemented using ultra-wideband (UWB)~\cite{van2020board}\cite{niculescu2022iotj} or other wireless communication protocols~\cite{liu2007tos}\cite{coppola2018ar}.
To tackle the sensing and compute constraints on nano-UAVs, off-board processing was proposed for pose estimation~\cite{simsek2021blur}\cite{micheal2010ram}. However, these methods are unsuitable for many application scenarios, as they require prior installation of external infrastructure and reliable communication between the mobile agent and the base stations. 

\begin{figure}[t]
  \centering
  \includegraphics[width=0.98\linewidth]{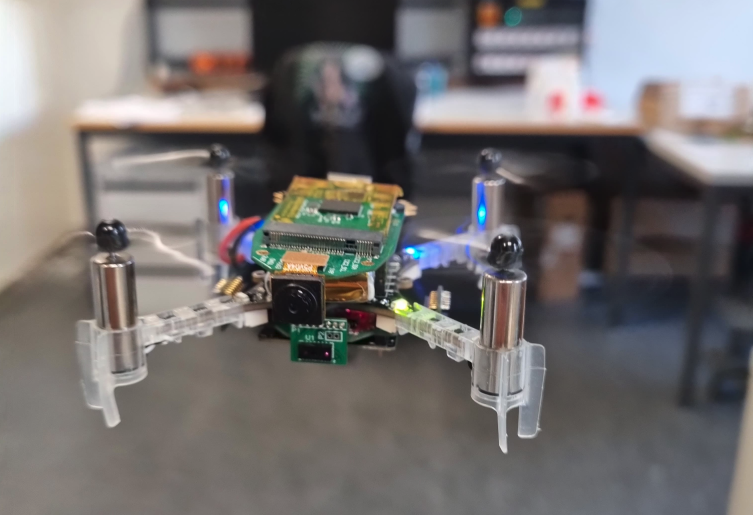}
  \includegraphics[angle=0,width=0.492\linewidth]{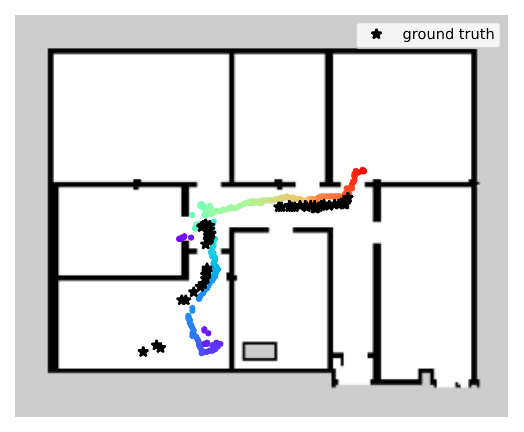} 
   \includegraphics[angle=0,width=0.492\linewidth]{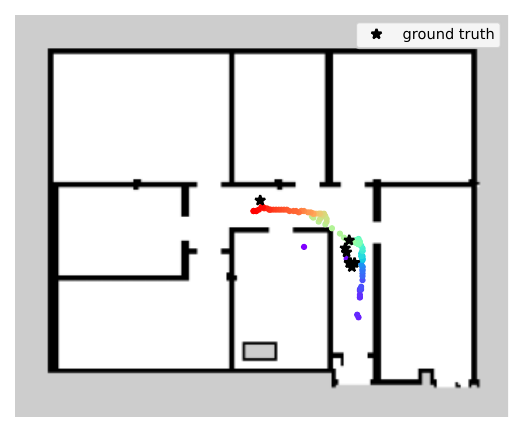} 
  \caption{Top: A nano-UAV while flying and globally localizing in an office environment using our novel sensor fusion approach. Bottom: A qualitative evaluation of the localization results on recorded sequences. Ground truth pose is marked by black stars. The rainbow colors encode the time of prediction, with purple marking the beginning of the sequence and red its end.}
  \label{fig:motivation}
\end{figure}
\vspace{-0.1cm}
Map-based approaches for localization do not require pre-existing infrastructure or external localization cues, and can therefore operate independently in indoor environments even when communication is unavailable. In map-based approaches, the agent uses measurements acquired by its sensors (LiDAR, camera, sonar, etc.) to estimate its pose in a given map, such as a landmark-based map~\cite{betke1997tra}\cite{gutman2002icpr} or an occupancy grid map~\cite{moravec1989sdsr}\cite{elfes1989phd}. 

Range sensors have been successfully coupled with occupancy grid maps~\cite{dellaert1999icra}\cite{thrun2001ai} for indoor localization. Unfortunately, these sensors are power-hungry and large and therefore unsuitable for use on a nano-UAV~(\figref{fig:motivation}), where only around 10\% (around 1-2 Watts) of the overall power budget can be spent on sensing and processing without affecting the flight time substantially~\cite{elkunchwar2021toward}. Miniaturized time-of-flight (ToF) sensors~\cite{mueller2023date} were used to localize a nano-UAV. However, they suffer from a short range (3m) and low beam count ($8\times8$) that limits their ability to operate in large spaces.

Sparse maps, such as floor plans, are attractive due to their availability, removing the need for a complex mapping procedure before deployment. However, relying solely on geometric information from range sensors can lead to global localization failures in sparse maps and environments with high structural symmetries~\cite{zimmerman2023ral}. 

Humans navigate using objects rather than precise metric measurements~\cite{mendez2018icra}\cite{yi2009iros}, which inspires leveraging semantic information to improve localization. With the recent advances in tasks such as object detection~\cite{bochkovskiy2020arxiv}, semantic cues are commonly utilized for robot localization~\cite{atanasov2015ijrr}\cite{yi2009iros}\cite{zimmerman2023ral}. However, they still require high-end, energy-consuming computational resources, which are usually not available on nano-UAVs. The challenge of executing semantic inference under limited computational resources is addressed with neural architecture search, quantization and optimized deployment engines~\cite{fedorov2019nips}\cite{nni2021github}\cite{warden2019oreilly}.  


Our main contribution is an approach for global indoor localization on a nano-UAV. In our approach, we use sensor fusion to exploit both semantic and geometric information. We fully execute our online and onboard approach on a novel low-power processor. We address the difficulty of executing object detection tasks on resource-constrained platforms, describing a pipeline for model quantization and deployment. We introduce a memory-efficient map representation which contains both semantic and geometric information. We utilize semantic information extracted from the camera and fuse it with range measurements from a miniaturized ToF sensor to localize a nano-UAV in a the metric-semantic map, by introducing a novel observation model, as seen in \figref{fig:motivation}. 



In our evaluation, we demonstrate that our approach can 
(i) globally localize a nano-UAV in a given map,
(ii) infer semantic information under resource constraints,
(iii) operate with low power consumption onboard
(iv) execute in real-time (15Hz ToF, 5Hz camera). Additionally, we provide a video of the live demo, as well as an open-source implementation. 

\vspace{-0.25cm}
\section{Related Work}
\label{sec:related}


Localization is a fundamental problem in robotics and has been extensively-researched~\cite{cadena2016tro}\cite{thrun2005probrobbook}. Localization is often cast within a probabilistic framework, accounting for the uncertainty of sensor measurements, and providing accurate and robust state estimation. The foundational works of probabilistic robot localization include Markov localization by Fox \etalcite{fox1999jair}, extended Kalman filter~\cite{leonard1991tra} and particle filter-based localization, commonly known as Monte Carlo localization (MCL) by Dellaert \etalcite{dellaert1999icra}. These works focused on localization using range sensors such as 2D LiDARs and sonars, and later works also utilized cameras~\cite{bennewitz2006euros}. 

Localization in indoor human-oriented environments is particularly challenging, due to the presence of dynamic obstacles such as humans, chairs and carts, as well as quasi-static changes such as opening and closing of doors and rearrangement of furniture~\cite{zimmerman2023iros}. Relying solely on geometric features may lead to global localization failure, especially when localizing on sparse maps such as floor plans~\cite{zimmerman2023ral}. 

Additional sources of information, such as WiFi and textual cues, have been integrated into localization frameworks~\cite{cui2021iros}\cite{ito2014icra}\cite{zimmerman2022iros} to increase the robustness of localization, as well as semantic information about objects in the scene. Mendez \etalcite{mendez2018icra} exploit semantic cues to localize, but only address a small set of semantic classes (walls, windows and doors) which can lead to global localization failure in the case of structural symmetries. In our previous work~\cite{zimmerman2023ral}, we propose a semantic localization utilizing both 2D laser and a camera. While our current approach shares the concept of abstract semantic map representation, the previous method requires a $360^o$  laser and camera coverage, and a power-hungry onboard computer~(Intel NUC10). 

In the context of localization for small UAVs, the most commonly-used approaches rely on preexisting infrastructure. Coppola \etalcite{coppola2018ar} present a Bluetooth based relative localization approach where signal strength is used as a range measurement. Van \etalcite{van2020board} propose UWB for both communication and relative localization of micro-UAVs. Similarly, Niculescu \etalcite{niculescu2022iotj} suggest a global localization approach based on UWB for nano-UAVs. Unlike our map-based approach, these methods require the presence of pre-installed infrastructure and reliable communication between the UAVs and the base stations. 
In our previous work~\cite{mueller2023date} we introduce a map-based global localization approach relying on miniaturized ToF sensors, which can be executed online onboard a nano-UAV. While the approach shows sufficient accuracy in very narrow and cluttered environments, it cannot operate in larger, open environments we target in our work due to the short range of the ToF sensors~(3m).

Extracting semantic information is essential for semantically-guided localization. While lightweight architectures such as YOLO~\cite{bochkovskiy2020arxiv} enable inference of object detection models onboard computers such as Intel NUC and Nvidia Jetson, they still require several tens of megabytes of memory for sensor-rate execution and are therefore unsuitable for execution on microcontrollers(MCU). Recent years witness a growing interest in the deployment of machine learning on edge devices, specifically semantic perception tasks such as object detection~\cite{edge2022aivision}\cite{lin2020nips}\cite{moosmann2023arxiv}.  Motivated by these trends, our approach incorporates also semantic information from the onboard camera, to overcome the range limitation of the ToF sensors. 

To the best of our knowledge, our approach is the first to fuse both range measurements and semantic cues for global localization on nano-UAVs. By optimizing the map format and sensor model for a novel ultra-low-power processor, we are able to globally localize, onboard and online, in indoor environments without the need for infrastructure or reliable communication~\cite{niculescu2022iotj,van2020board}. Additionally, we are the first to deploy a state-of-the-art~(SotA) object detection model from the YOLO family on a RISC-V multi-core platform, achieving 20fps with only 2.5mJ per frame.




\section{System Overview}
\begin{figure}[t]
  \centering
   \includegraphics[angle=270,width=0.99\linewidth]{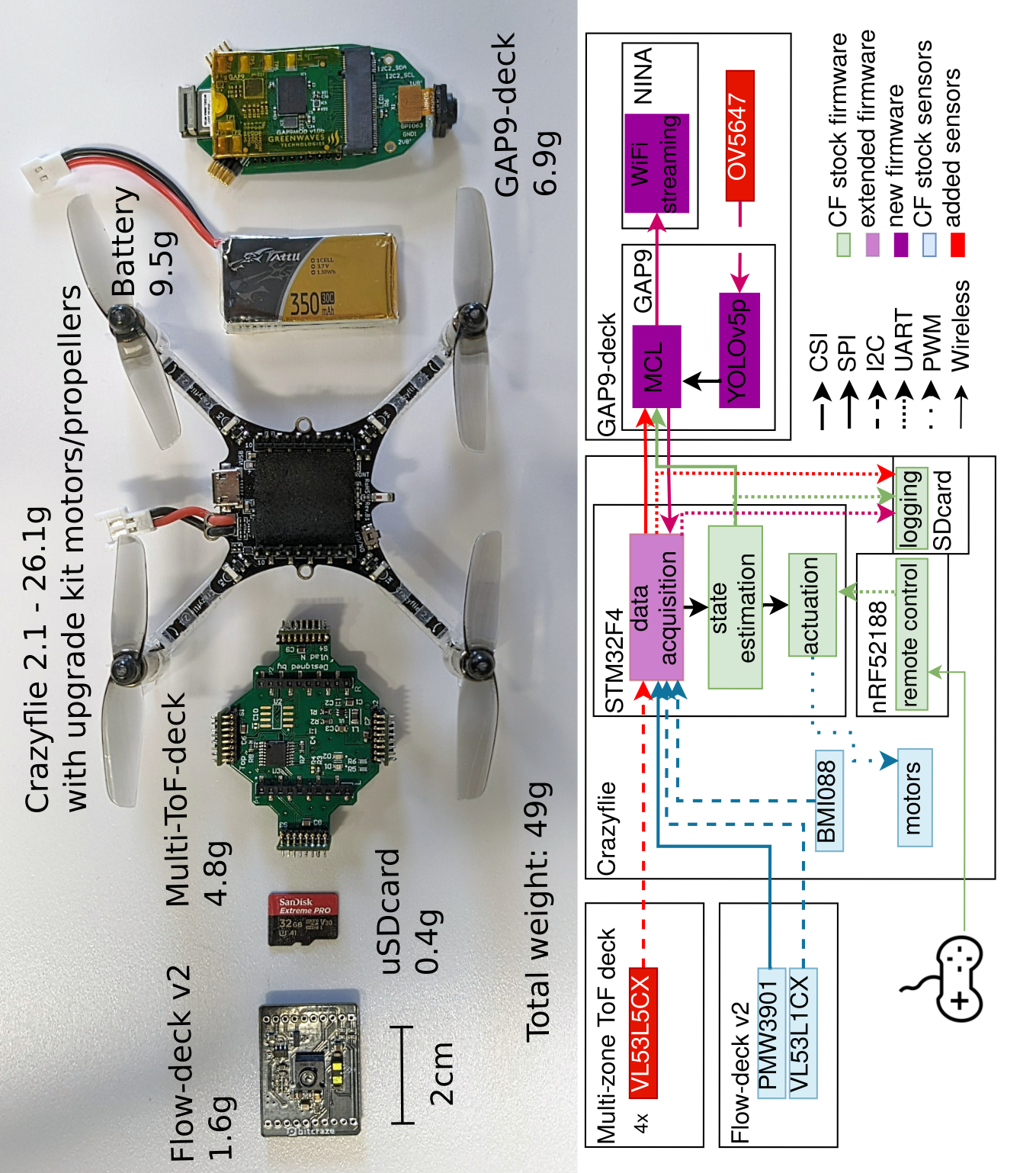}
  \caption{System overview. Top: All stacked components on the nano-UAV, ordered from the top (left) to the bottom (right). Bottom: A visualization of the communication paths and task distribution between all employed processors and sensors.}
  \label{fig:hw}
\end{figure}

We introduce the hardware setup including the nano-UAV, sensors, and compute platform.
We use a Crazyflie 2.1, an open-source drone, and extend it with three plug-on decks, as shown in \figref{fig:hw}. 

We equip the Crazyflie 2.1 with upgrade-kit motors and propellers and a 350mAh battery. The Crazyflie already features two MCUs (STM32F405 and nRF52188) and an inertial measurement unit. For improved state estimation we mount a flow-deck v2 that provides visual odometry with a downward-facing optical flow and ToF sensor.
To enable semantic localization, we mount miniaturized ToF sensors, a tiny, low-weight camera and a novel, low-power multi-core processor on custom designed decks. 
The multi-zone ToF deck features 4 ToF sensors, each one detecting a 64-pixel grid with a 67 degree diagonal field of view (FoV)~\cite{friess2023arxiv}. Additionally, it includes an uSDcard for logging.
The GAP9-deck, as introduced in~\cite{mueller2023date}, integrates an RGB camera (OV5647) and a RISC-V parallel system-on-chip called GAP9\footnote{\href{https://greenwaves-technologies.com/gap9_processor}{https://greenwaves-technologies.com/gap9\_processor}} as processing platform. It also includes a NINA WiFi module, which is used for data collection.



Our approach requires a camera interface and the execution of multiple computationally heavy loads with relatively high memory requirements ($>$1MB), making GAP9 a good fit. GAP9 is based on the open-source chip Vega~\cite{rossi2021vega} and features one RISC-V core as fabric controller (FC) and a cluster with one orchestrator and 8 worker cores. It features interleaved RAM (referred to as L2) shared between the cores, with memory capacity of 1.5MB. The maximum operating frequency of the cores is 370MHz for both the cluster and the FC. GAP9 also features NE16, a convolutional neural network~(CNN) hardware accelerator, specialized for highly efficient MAC operations. NE16 is tailored to 3$\times$3 convolutions, as it features 9$\times$9$\times$16 8x1bit MAC units, but it also offers support for 1$\times$1 and 3$\times$3 depth-wise convolutions and fully connected layers. Additionally to the aforementioned internal memory on GAP9, we mount an L3 RAM octa-SPI memory of 32MB. 

\figref{fig:hw} displays the interactions between the different components.   
Note that the WiFi module, the 2.4GHz radio and the uSDcard solely serve for logging and remote steering purposes, no computations are offloaded. For time synchronization of the logging we send a timestamp packet from the STM32 on the Crazyflie to the GAP9 every 10ms.
\vspace{-0.1cm}

\section{Approach}

We aim to globally localize a nano-UAV in a given floor plan, targeting full-scale human oriented environments, such as offices. To this end, we fuse geometric and semantic information, in a MCL framework. In \secref{sec:deploy} we detail the training and deployment pipeline for porting a SotA object detection model to a resource-constrained MCU. We briefly explain MCL~(\secref{sec:MCL}). In \secref{sec:semmap} we describe our optimized semantic map format, and then present our novel fusion sensor model in \secref{sec:fusion}.
\subsection{Object Detection}\label{sec:deploy}
We use a modified architecture from the YOLOv5 family, we refer to as YOLOv5p, to reduce the memory and execution time required for inference fully onboard, on the parallel RISC-V processor. The proposed YOLOv5p has a smaller backbone and also a reduced head, with 623K parameters compared to the 1.9M parameters of smallest YOLOv5 model(YOLOv5n). We first pre-train our model on the COCO dataset~\cite{lin2014eccv} for 100 epochs, and then fine-tune on our custom dataset, learning semantic classes of interest. To deploy the model on the GAP9, we use the deployment pipeline NNTool\footnote{\href{https://github.com/GreenWaves-Technologies/gap\_sdk/tree/master/tools/nntool}{https://github.com/GreenWaves-Technologies/\ gap\_sdk/tree/master/tools/nntool}}, which includes quantization, an inference engine for verification in python and code generation for deployment on resource-constraint MCUs. It supports the NE16, a hardware accelerator present in GAP9. 
\subsection{Monte Carlo Localization}\label{sec:MCL}
MCL~\cite{dellaert1999icra} is a probabilistic framework for estimating a robot's state. A particle filter is used to represent the robot's belief $p(\q{x_t}\mid \q{z_{1:t}}, m)$, where $m$ is a given map, $\q{z_t}$ are sensor measurements and $\q{x_t}$ is the robot's state.
Each particle $s_t^{(i)}\in \mathcal{S}$ contains the robot's state $\q{x_t}^{(i)}$ and its weight $w_t^{(i)}$. As we localize in a 2D grid, the state is given by $\q{x_t}^{(i)}=(x,y, \theta)^\top$. The proposal distribution is sampled when a control command $\q{u_t}$ is available, with odometry noise  $\sigma_{\text{odom}} \in \mathbb{R}^3$. A particle is re-weighted according to the sensor model, given a sensor reading $\q{z_t}$. In our MCL implementation, we perform the updates asynchronously whenever an odometry input or an observation is available, provided that the nano-UAV moved a threshold distance of $d_{xy}$ or rotated by more than $d_{\theta}$. 
\subsection{Semantic Map Format}\label{sec:semmap}
The proposed semantic map format contains geometric information, in the form of an occupancy grid map extracted from a floor plan, which is enhanced with prior semantic information. Similarly to our previous work~\cite{zimmerman2023ral}, we choose a simplified representation for our semantic information, defining objects by their semantic class and a 2D bounding box. However, the contribution of our work is a unified, memory-efficient map representation, instead of multiple semantic visibility maps. In the proposed approach, the occupancy states (free, occupied, unknown) are represented by 2 bits, and the semantic maps are represented by 1 bit per class, resulting in one 16-bit map. This generic representation of semantic objects enables fast, manual annotation and removes the need for an expensive mapping procedure. Our approach can handle inaccurate annotations of both object pose and size. A colored visualization of the semantic maps can be seen in \figref{fig:floorplan}.
\subsection{Geometric-Semantic Fusion Sensor Model}\label{sec:fusion}

The output of our object detection model contains the class label, the bounding boxes coordinates in a $xyxy$ format and the confidence score for the detected object. First, we compute the center of the bounding box $\q{v_c}=(x_c, y_c, 1)$ in homogeneous coordinates and project it to a 3D ray in the camera frame 
 \begin{align}
\q{V_c}(\lambda) = O + \lambda\mq{R}^{-1}\mq{K}^{-1}\q{v_c},
\end{align}
where $\mq{K} \in \mathbb{R}^{3 \times 3}$ is the camera calibration matrix and $O\in \mathbb{R}^3$ is camera center. As the camera is aligned to the forward direction of the nano-UAV, we assume the rotation matrix is unity. For every particle $s_t^{(i)}\in \mathcal{S}$, we transform the 3D ray $\q{V_c}(\lambda)$ to the map coordinate frame using the pose $\q{x_t}^{(i)}$. Then we trace the ray in the semantic map, from $\q{x_t}^{(i)}$ in the direction of the transformed ray $\q{V_c^m}(\lambda)$. We consider the tracing to fail if we ran into a wall in the occupancy grid map before reaching an object of class $c$, and in this case we penalize the particle by lowering its weight. As the FoV of the front ToF sensor and the camera overlap, we can associate range measurements to detected objects. If the tracing was successful, we look up the $8\times8$ ToF beam measurements corresponding to the bounding box, and calculated the average distance to the object $d_{\text{ToF}}$. If the distance is less than $\tau_{t}$, we match the traced distance $d_{\text{trace}}$ to the measured distance $d_{\text{ToF}}$
 \begin{align}
 p_s(\q{z_t}\mid m, \q{x_t}) = \frac{1}{\sqrt{2 \pi \sigma_{\text{s}}}}\exp{\left(-\frac{(d_{\text{trace}} - d_{\text{ToF}})^2}{2\sigma_{\text{s}}^2}\right)},
\end{align}
where $\q{z_t}$ includes both the semantic observation inferred by our object detection model and range measurements from the front ToF sensor. When semantic information is not available, we use the ToF observations to re-weight the particles according to the Beam End Model~\cite{thrun2005probrobbook}
\begin{align}
 p_g(\q{z_t}\mid m, \q{x_t}) &=  \frac{1}{\sqrt{2 \pi \sigma_{\text{g}}}} \exp{\left(-\frac{\mathrm{edt}(\q{\hat{z}_t)}^2}{2 \sigma_{\text{g}}^2}\right)},
 \label{eq:beam_end_model}
\end{align}
where $\q{z_t}$ includes 32 range measurements, extracting 8 beam points from the middle row of our 4 ToF sensors. $\q{\hat{z}_t}$ refers to the end points of the beams in the occupancy grid map, after considering the particle's pose. EDT is the Euclidean distance transform~\cite{felzenszwalb2012toc}, which was truncated at distance $r_{\text{max}}$ and quantized to 8-bit for memory efficiency. As the ToF measurements are unreliable after 3\,m, we discard observations beyond that range, and only perform the update step with enough valid measurements. We employed an aggressive resampling strategy, sampling the particle after every observation.





\section{Experimental Evaluation}
\label{sec:exp}

%
%
We evaluate the proposed method in several experiments, to support our claims the we can (i) globally localize a nano-UAV in a given map, (ii) infer semantic information under resource constraints, (iii) operate with low power consumption onboard, (iv) execute in real-time (15Hz ToF, 5Hz camera). Specifically, we show that we can localize in a featureless map such as a floor plan using a low-power compute platform and miniaturized sensors, due to leveraging semantic information.

\subsection{Experimental Setup}

\begin{figure}[t]
  \centering
  \includegraphics[angle=0,width=0.49\linewidth]{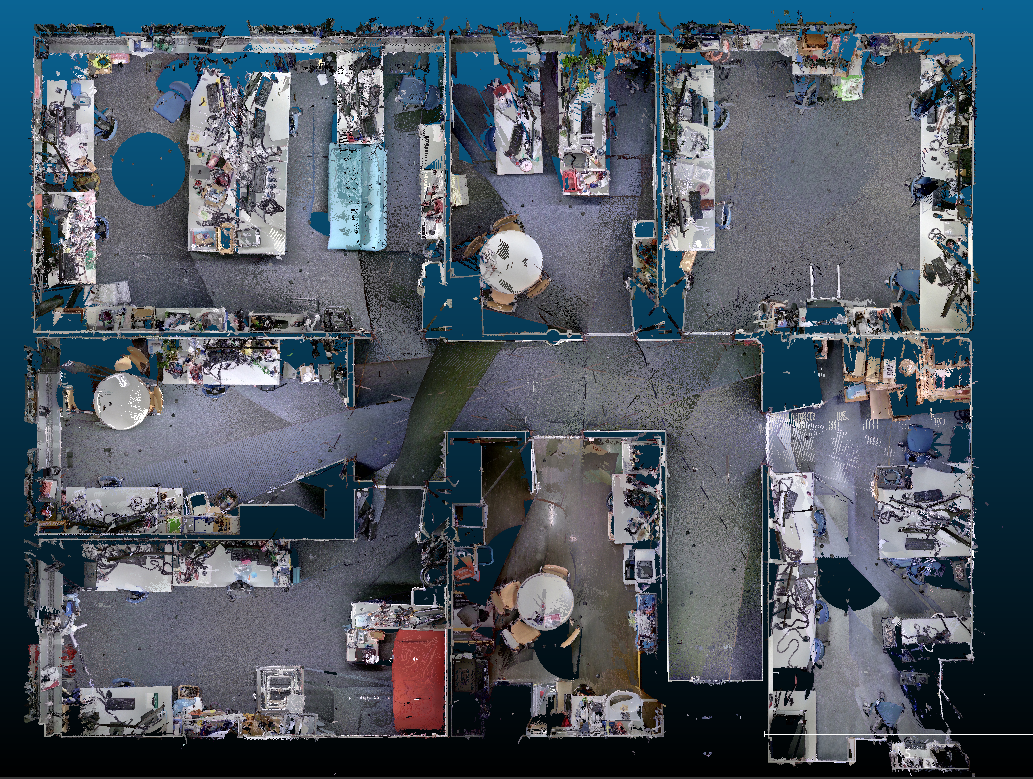}
 \includegraphics[width=0.47\linewidth]{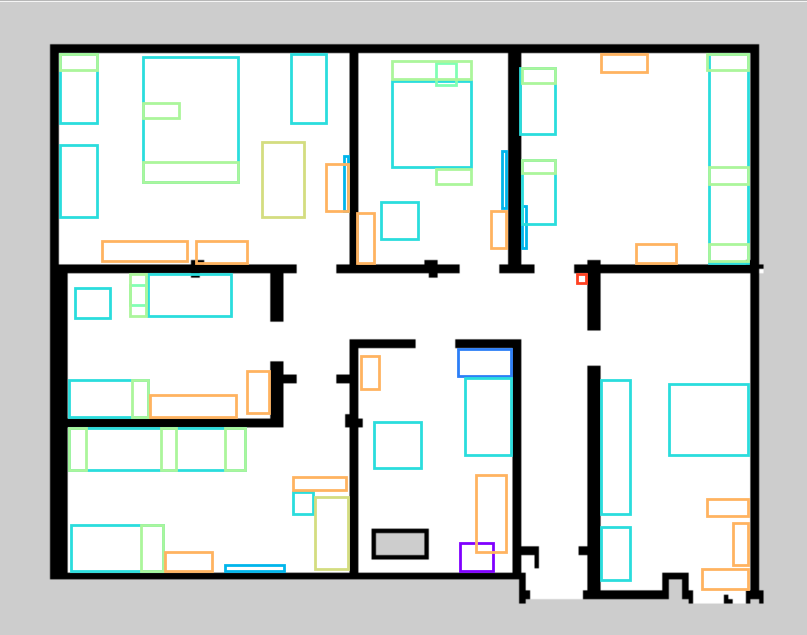}
  \caption{Left: A top view of the dense pointcloud captured with the Z+F Imager 5016 terrestrial laser scanner, which was used solely for GT extraction. The full pointcloud has 200 million points. Right: The semantically-enriched floor plan of the lab. Semantic objects of interest are represented using their bounding box and class ID. Different colors represent different object classes. The semantic information was added manually, without a complex measuring or mapping procedure.}
  \label{fig:floorplan}
\end{figure}

To evaluate our approach, we recorded 10 drone flights (S1-S10) over several weeks, spanning across the lab~(\figref{fig:floorplan}), including quasi-static changes, furniture moving and different lighting conditions. The recordings include odometry from the Crazyflie's internal state estimation and ToF measurements at 15Hz. For the front ToF sensor, we recorded the full $8\times8$ grid, while for the right, back and left ToF sensors, we only recorded 8 beams extracted from the middle row of the grid, due to bandwidth limitations. We also recorded images with $640\times480$ pixel at a lower rate of 2Hz, due to Wi-Fi streaming limitations. 

To assess the accuracy and robustness of our approach, we used AprilTags~\cite{olson2011icra-aara} to compute the pose of the nano-UAV. We placed 148 AprilTags on the walls, and then used a SotA laser scanner to create a dense 3D pointcloud of the lab, shown in~\figref{fig:floorplan}. The 3D coordinates of the AprilTags were extracted by running the detector on orthographic projections of the walls. In images where the AprilTags are detectable, we used 2D-3D correspondences to estimate the nano-UAV's position~\cite{marchand2015pose}. The images were recorded at $640\times480$ pixel to enable the detection of AprilTags. Due to the low resolution and the blurry nature of in-flight images, the GT accuracy is $\sim 0.1m$. The AprilTags are used strictly for evaluation and are not part of the localization approach. In addition, we collected training, validation and test sets to train and benchmark our object detection model.

The given map~(\figref{fig:floorplan}) is a floor plan augmented with semantic information in the form of bounding boxes, annotated by a lab member from memory. The semantic annotations are not precise in position or size, and did not require any type of measurement, but are simply hand-drawn. The map resolution is 0.05m/pixel, covering an area of 280~$\text{m}^2$. 

\vspace{-0.1cm}





\subsection{Object Detection Performance}
\begin{figure}[t]
  \centering
  \includegraphics[angle=0,width=0.48\linewidth]{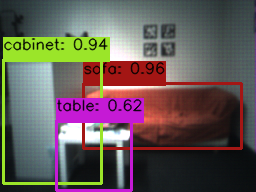} 
   \includegraphics[angle=0,width=0.48\linewidth]{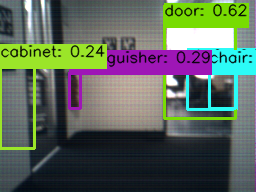} 
  \caption{A qualitative evaluation of the 8-bit quantized object detection model on $256\times192$  input images.}
  \label{fig:qual_detection}
\end{figure}

We evaluated the quantized YOLOv5p object detection model to ensure that our deployment process can preserve the accuracy of the full precision model. While images are acquired at 640x480 pixel, they are downsampled to $256\times192$ for inference. We compared the average precision for each class of interest and the mean average precision for 4 variations. We trained YOLOv5p using the Ultralytics framework~\cite{jocher2023github}. FP32 is a full precision model converted by the NNTool from YOLOv5p. MIXED is a quantized model with varying precision - the first layer is FP16, and the rest are UINT8. UINT8 is a NNTool 8-bit quantization of YOLOv5p. YOLOv5p inference was executed on a NVidia GTX3070 GPU. MIXED and UINT8 inferences were executed on the GAP9. 
For the evaluation, we collected a test set with 50 images, including several instances of each class of interest. The images were captured by the nano-UAV's onboard camera. As can be seen in \tabref{tab:map_score}, we lose up to 7.2\% accuracy in the quantization and deployment process, but the performance is still satisfactory, enabling the extraction of semantic information for localization. Inference examples from the UINT8 model can be seen in \figref{fig:qual_detection}.

Our object detection pipeline consists out of four parts: (i) image acquisition (ii) preprocessing (iii) quantized neural network (iv) post-processing. 
As we experienced limitations in the camera acquisition speed it takes 50ms to acquire an image. As the image is acquired by the $\mu$DMA~\cite{rossi2021vega}, with double buffering this time can be used productively on GAP9. The second part is preprocessing, which includes demosaicing and transforming from 10 to 8 bit inputs and takes 5ms. The quantized neural network takes 38ms, and the post-processing (non-maximum suppression) takes 0.3ms. This means that the limiting factor is the image acquisition - currently we can reach 20fps, however, even if the hardware would allow to acquire images faster 23fps would still be the upper limit for the UINT8 network.
\vspace{-0.2cm}
\begin{table}[t] 
  \caption{Average precision(AP) (IoU=0.50) scores for the test set, confidence TH 0.2, IoU TH 0.5}
  \centering 
  \resizebox{\columnwidth}{!}{
\begin{tabular}{ccccccccccccc}\toprule

Class        & sink & door & fridge & board & table &  plant & drawers & sofa & cabinet & extinguisher & all\\ \midrule
YOLOv5p    & 0.663  & 0.579  & 0.952  & 0.574  & 0.51  & 0.379  & 0.604  & 1.0  & 0.826  & 0.967  & 0.705  \\
FP32     & 0.663 & 0.508 & 1.0 & 0.525 & 0.515 & 0.337 & 0.659 & 1.0 & 0.723 & 1.0 & 0.693 \\
MIXED  & 0.663  & 0.482  & 1.0  & 0.752  & 0.516  & 0.168  & 0.554  & 1.0  & 0.715  & 1.0  & 0.685  \\
UINT8   & 0.663  & 0.492  & 1.0  & 0.644  & 0.436  & 0.168  & 0.604  & 1.0  & 0.68  & 0.851  & 0.654  \\

\bottomrule 
\end{tabular}
}
\label{tab:map_score} 
\end{table} 
\vspace{-0.1cm}

\subsection{Global Localization in Floor Plans}
To evaluate the capability of our localization approach, we examined 3 metrics. The success rate, convergence time and absolute trajectory error~(ATE). Since our ground truth~(GT) positions are not continuous, as AprilTags are not visible or detectable in every frame, we only evaluated our predictions at the timestamps where the GT checkpoints were available. 
We consider the point of convergence to be when the ATE is lower than 0.5m. We consider a localization to be successful when the pose estimation remains converged until the end of the sequence. The algorithm parameters are specified in \tabref{tab:parameters}. We tested our approach (Nano-SMCL) using a particle filter with 4096 particles, that were initialized uniformly all over the map. For inferring the semantic cues, we used the 8-bit quantized model, with $256\times192$  input images.

\begin{figure}[t]
  \centering
  \includegraphics[angle=0,width=0.7\linewidth]{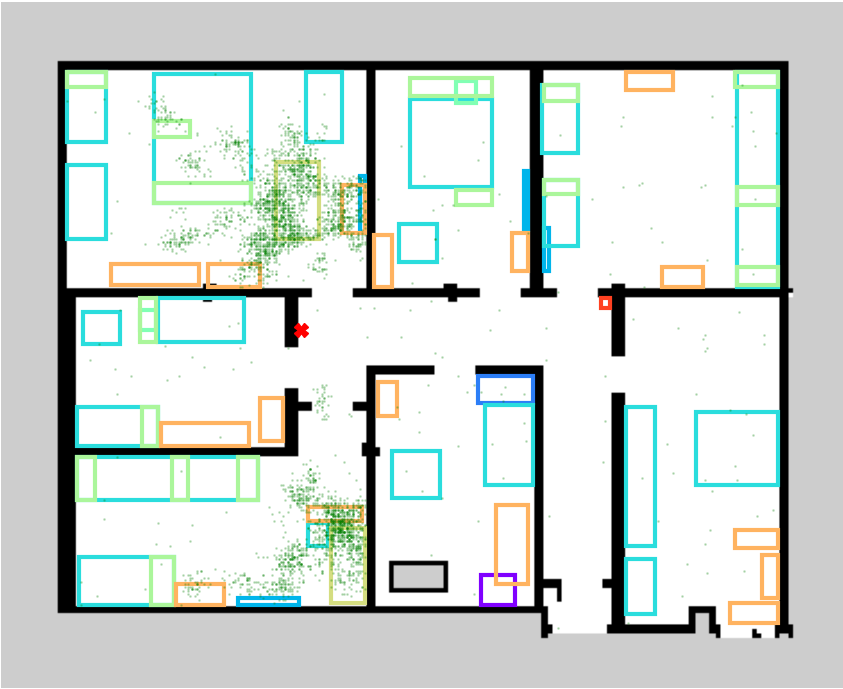} 
  \caption{A failed localization scenario due to ambiguity in both geometric and semantic features. The particles, marked as green dots, are divided between two rooms with similar properties. The weighted-average prediction is marked with a red cross.}
  \label{fig:qual_failure}
\end{figure}

As a baseline, we used the ToF-based MCL approach~\cite{mueller2023date}, referred to as Nano-MCL, providing input from 4 ToF sensors and not relying on semantics. As portrayed in \tabref{tab:ate_4096}, our algorithm converges with $90\%$ success rate, with an average ATE of 0.32m. Our average convergence time is 45s. A qualitative evaluation of our localization results can be seen in \figref{fig:motivation}. Our approach failed to localize on sequence S7. In this short sequence, the nano-UAV was mostly in the center of a large and cluttered room, and we could not make use of the ToF measurement. In addition, there is also an ambiguity related to the semantic information, where the configuration of sofa, cabinet and table in close proximity also exists in another room, causing the particle filter to maintain two hypotheses as can be see in \figref{fig:qual_failure}. We outperformed the range-only MCL approach on all criteria, showing that relying solely on geometric information leads to a success rate of only 10\%  We speculate that Nano-MCL is suitable for more detailed maps such as occupancy grid maps, and also for environments without vast empty spaces. This is a limiting factor for the short-sighted ToF sensor, and motivates the use of both semantic and geometric information.

\begin{table}[t]
  \caption{Algorithm parameters}
   \centering
   \resizebox{\columnwidth}{!}{
 \begin{tabular}{cccccccc}\toprule
 Method & $\sigma_{\text{odom}}$  &  $\sigma_{\text{g}}$ & $\sigma_{\text{s}}$ & $ \tau_{t} $ & $r_{\text{max}}$ & $d_{\text{xy}}$ & $d_{\theta}$\\ \midrule
 Nano-SMCL & (0.5\,m, 0.5\,m, 0.5\,rad) & 8.0 & 10.0 & 2.5\,m & 2\,m & 0.05\,m & 0.05\,rad\\
 Nano-MCL & (0.5\,m, 0.5\,m, 0.5\,rad) & 8.0 & - & - & 2\,m & 0.05\,m & 0.05\,rad\\
  \bottomrule
 \end{tabular}
 }
 \label{tab:parameters}
\end{table}
\vspace{-0.1cm}

  \begin{table}[t] 
  \caption{Evaluation of the approaches on recordings S1-S10 with 4096 particles. Top: Absolute trajectory error in meters. Bottom: convergence time in seconds.}
  \centering 
  \resizebox{\columnwidth}{!}{
\begin{tabular}{ccccccccccccc}\toprule

Method        & S1 & S2 & S3 & S4 & S5 & S6 & S7 &S8 & S9 & S10 & AVG \\ \midrule
Nano-SMCL & 0.47  & -  & 0.30  & 0.22  & 0.36  & 0.22  & 0.40  & 0.25  & 0.29  & 0.37  & 0.32 \\
Nano-MCL & -  & -  & - & -  & -  & 0.26  & -  & -  & -  & - & 0.26 \\
\midrule 
Nano-SMCL & 30.87  &  -  & 42.39  & 101.43  & 19.30  & 30.32  & 56.04  & 54.78  & 24.69  & 46.57  &  45.15 \\
Nano-MCL     &  -  & - &  -  &  -  &  -  & 67.02  &  -  &  -  &  -  & - &  67.02 \\
\bottomrule 
\end{tabular}
}
\label{tab:ate_4096}
\vspace{-0.1cm}
\end{table} 
\vspace{-0.1cm}


\vspace{-0.1cm}




\subsection{Real-time execution, power and memory footprint}
In this section, we present the execution times, power measurements and memory footprint of our approach.

\textit{Execution times}
In Table \ref{tab:exec_times} we summarize the average execution times (FC and cluster running at 370MHz) of the different steps, and the resulting processor load per task at the worst-case execution rate, using the maximal 15Hz from the ToF sensor for the ToF and odometry update, as well as 5fps for the camera (all tests are executed at around 2fps, limited by streaming for debugging and repeatability purposes).

When parallelizing on 8 cores, we reach an overall average speedup of 4.5 on the MCL, leading to a maximum worst-case load of 0.77 and demonstrating that our semantic localization approach can run in real-time. Note that this is a worst-case scenario for computations, since we assume the maximum frequencies we can acquire data with. In practice, the MCL updates are only performed under certain conditions, such as having valid observations and moving a threshold distance.  
Note that the camera acquisition time does not imply any processor load per se, as it can be executed by the $\mu$DMA~\cite{rossi2021vega}, however, due to camera driver limitations we can only acquire images with the full FoV in VGA resolution and with 10 bits per pixel, saved in 2 bytes, resulting in a too big image to double buffer it in L2. The data coming from the Crazyflie's STM32 (odometry estimation and ToF measurements) is much smaller and can be double buffered.
  \begin{table}[t] 
  \caption{Execution time, worst-case execution rate, and resulting processor load for single- and multi-core implementations (where present).}
  \resizebox{\columnwidth}{!}{
\begin{tabular}{cccccccccccc}\toprule

       & \multirow{2}{2em}{camera acqu.} & \multirow{2}{2em}{pre- process} & NN & \multirow{2}{2em}{post- process}. & \multirow{2}{3em}{cam/ToF fusion} & \multirow{2}{2em}{obs. ToF} & motion &  \multirow{2}{2em}{resampling} & \textbf{total}\\ \\ \midrule
Worst-case exec. rate  (Hz) &  5  & 5 &  5  & 5 & 15 & 15 & 15 & 15 & \\
Single-core (ms) &  50 & 4.5 &  -   & 0.3 & 43.9 & 39.1 & 10.3 & 0.6  & \\
Load single-core & 0.25 & 0.02 & -  & 0.00 & 0.66 & 0.58 & 0.15 & 0.01 & \textbf{1.67 + CNN} \\
Multi-core  (ms) &  -  & -  &  38  & - & 8.5 & 8.6 & 2.6 & 0.5 & \\
Load multi-core & (0.25) & (0.02) & 0.19  & (0.00) & 0.13 & 0.13 & 0.04 & 0.01 & \textbf{0.77} \\
\bottomrule 
\end{tabular}
}
\label{tab:exec_times} 
\end{table} 

\textit{Power consumption}
Power consumption of the drone can be divided to 3 main categories: actuation, sensing, and processing, with the motors consuming the most at $\sim 15W$.
The ToF sensors require 266mW each in continuous operation, accumulating to 1.06W, while the camera consumes only 90mW.
The Crazyflie stock electronics consume $\sim$280mW. The added processor, GAP9, consumes on average 64.7mW during the quantized YOLOv5p execution and 23mW during the MCL execution, which results in an average processing and sensing power consumption of just under 1.5W, staying inside the 10\% power budget usually advised for sensing and computation on nano-UAVs. For comparison, onboard compute platforms for standard-size robots, such as Intel NUC10, consume up to 120W, 5 orders of magnitude more than the GAP9.

\begin{figure}[t]
  \centering
  \includegraphics[angle=0,width=0.98\linewidth]{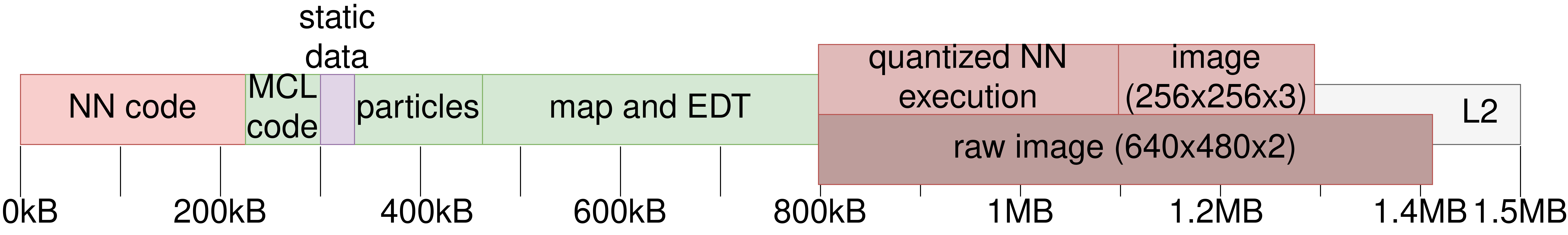} 
  \caption{The 1.5MB L2 memory on GAP9 is used for code and data.}
  \label{fig:L2}
\end{figure}

\textit{Memory}
The main constraint, fitting everything into the assigned L2 space, which we illustrate in~\figref{fig:L2} 
As we update the MCL at up to 15Hz, we allocate L2 memory for the particles for faster access. This requires 128kB for 4096 particles, where each particle's state $s_t^{(i)}= (\q{x_t^{(i)}}, w_t^{(i)})$ is represented by 4 floats (32-bit). For the 371x302 pixel semantic map, each pixel is represented by 16-bit value to encode the occupancy grid map and the multiple semantic maps. We also store a quantized 8-bit EDT, and both map and EDT require 336kB. We acquire raw images in VGA resolution ($640\times480\times2$2bytes), totaling in 614kB, and preprocess them to $256\times256$  8-bit RGB images, reducing the memory consumption to 196kB.
The NNTool allows to limit L2 size allocated for the model inference in exchange for slower execution (as transfers from external octa-SPI RAM are necessary). We allocated 300KB for our 8-bit quantized YOLOv5p as the inference time was only 3ms longer than with 1MB. Additionally, code and static data occupy L2 memory as well - 75KB for the MCL code, 225Kb for the neural network inference and 34kb for static data. The overall peak memory usage is 1.41MB out of the available 1.5MB, enabling the implementation of additional functionalities such as obstacle avoidance or navigation. 
\vspace{-0.1cm}
\vspace{-0.1cm}
%



\section{Conclusion} 
\label{sec:conclusion}
This paper presents a fully-onboard, global localization approach for a nano-UAV, operating in a full-scale, human-oriented indoor environment.  The proposed approach exploits low-element count, miniaturized ToF sensors,
fusing the range measurements with semantic information extracted from the onboard camera, to localize in a semantically-enhanced floor plan. We present a SotA object detection model at 20fps and 2.5mJ per frame on a $< 100$mW RISC-V multi-core processor.
We provide an optimized semantic map format and a sensor model that are suitable for onboard, online execution on nano-UAVs. In our experiments, we show the benefit of exploiting semantic cues for localization, and demonstrate that our approach can successfully localize in various real-world scenarios. 

\section{Acknowledgments}
We would like to thank Hilti for assisting with the 3D pointcloud acquisition. 



\bibliographystyle{plain_abbrv}

\bibliography{glorified,new}

\end{document}